\documentclass[10pt,twocolumn,letterpaper]{article}

\usepackage{wacv}
\usepackage{times}
\usepackage{epsfig}
\usepackage{graphicx}
\usepackage{amsmath}
\usepackage{amssymb}
\usepackage[small,labelfont=bf]{caption}
\usepackage{multirow}
\usepackage{subfig}

\wacvfinalcopy 

\ifwacvfinal\fi
\begin{document}

\title{Semi Supervised Phrase Localization in a Bidirectional Caption-Image Retrieval Framework}

\author{Deepan Das \hspace{1cm} Noor Mohammed Ghouse \hspace{1cm} Shashank Verma \hspace{1cm} Yin Li \\
University of Wisconsin-Madison\\
{\tt\small ddas27, mohamedghous, sverma28, yin.li@wisc.edu}
}

\maketitle
\ifwacvfinal\thispagestyle{empty}\fi
\begin{abstract}
We introduce  a novel deep neural network architecture that links visual regions to corresponding textual segments including phrases and words. To accomplish this task, our architecture makes use of the rich semantic information available in a joint embedding space of multi-modal data. From this joint embedding space, we extract the associative localization maps that develop naturally, without explicitly providing supervision during training for the localization task. The joint space is learned using a a bidirectional ranking objective that is optimized using a $N$-Pair loss formulation. This training mechanism demonstrates the idea that localization information is learned inherently while optimizing a Bidirectional Retrieval objective. The model's retrieval and localization performance is evaluated on MSCOCO and Flickr30K Entities datasets. 
This architecture outperforms the state of the art results in the semi-supervised phrase localization setting.
\end{abstract}



\begin{figure}[!h]
    \centering
    \subfloat[\textit{Input: A man pulling a heavy cart of bricks}]{
        \includegraphics[clip, width=\columnwidth]{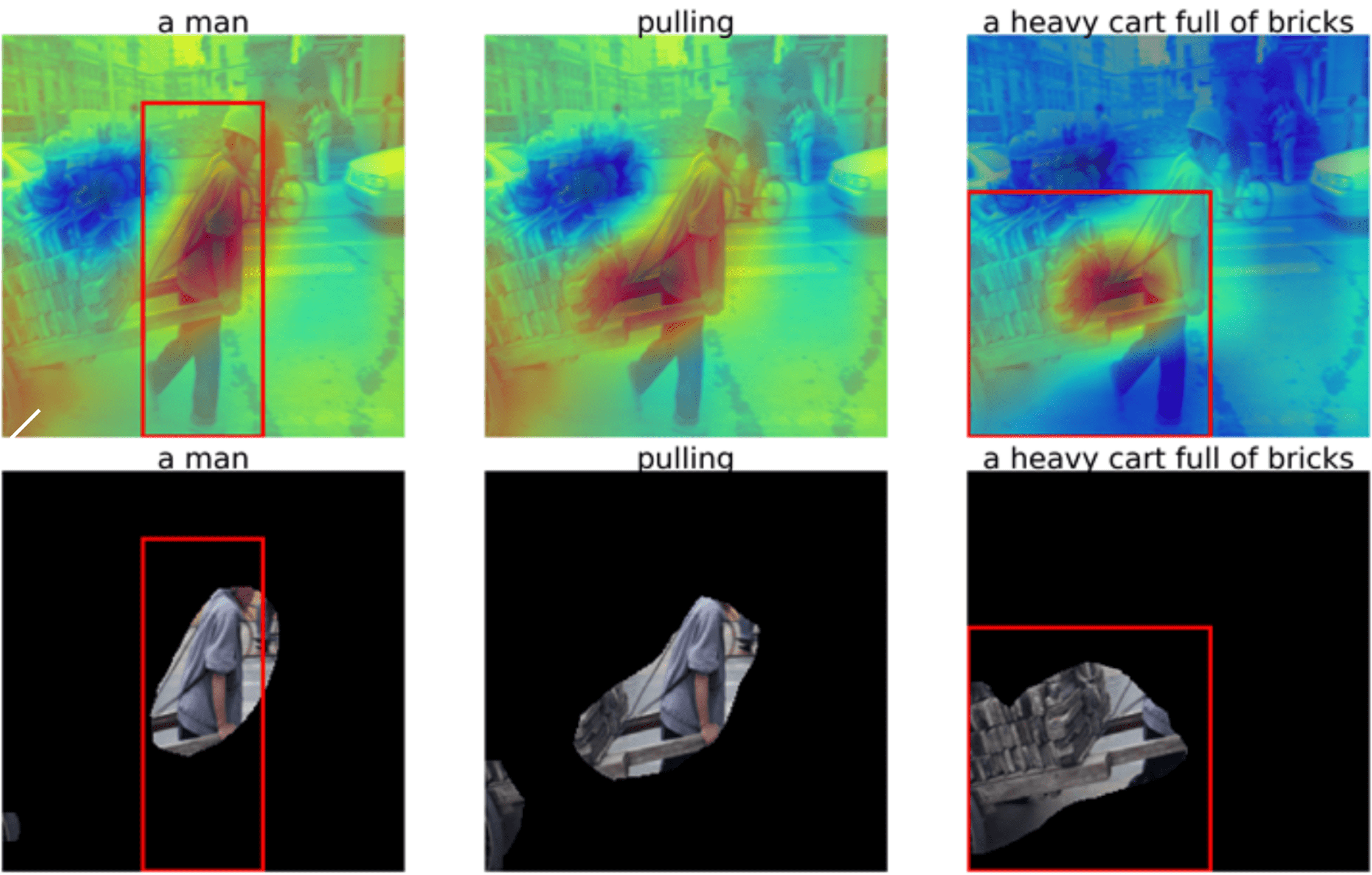}
    }\\
    \subfloat[\textit{Input: A large group of people are riding up and down on an escalator}]{
        \includegraphics[clip, width=\columnwidth]{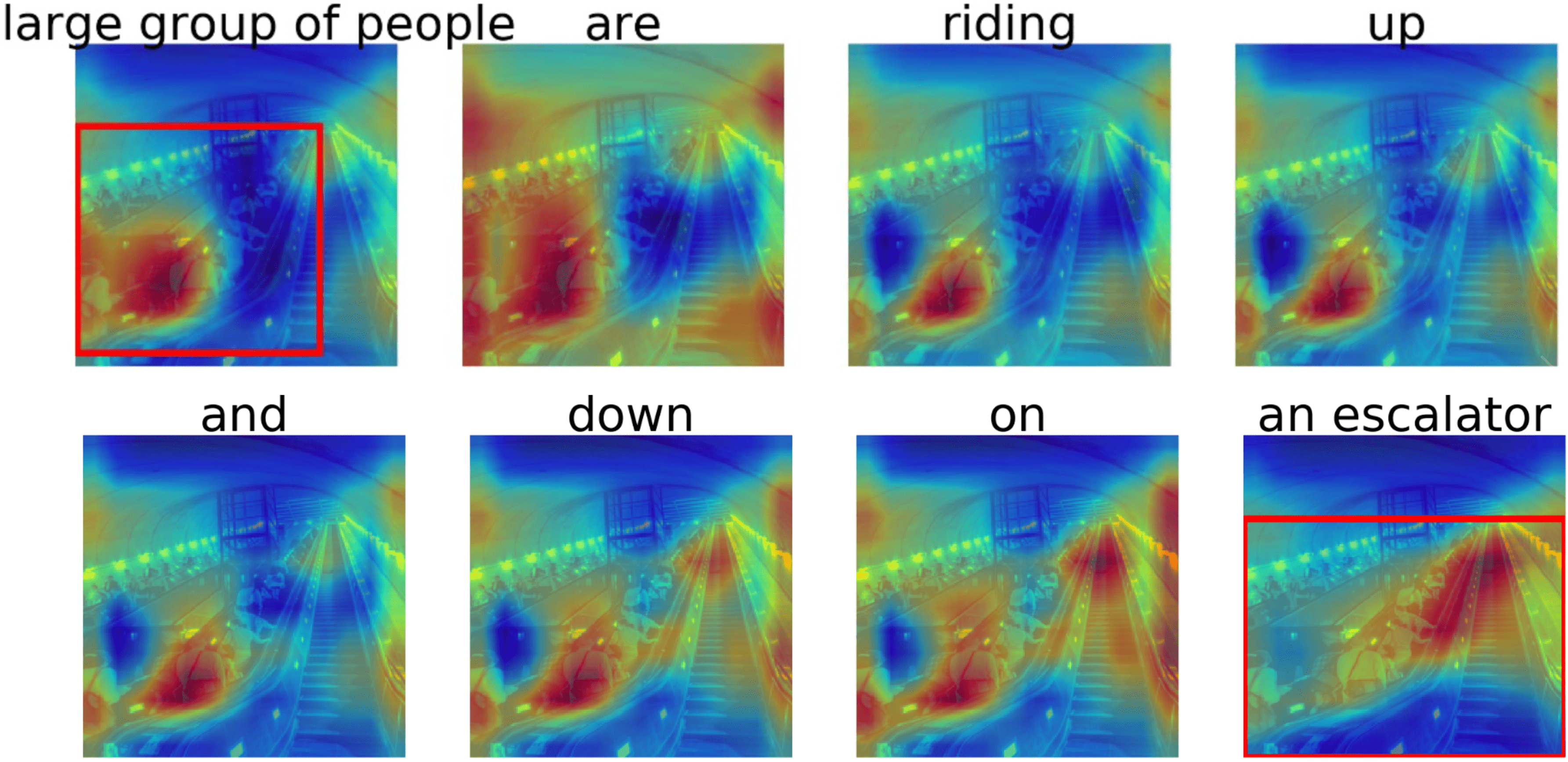}
    }
    \caption{We generate a spatial attention map for each constituent word. This method can also generate such maps for short phrases within the caption that describe a single entity. In (a) we have also generated segmentation masks from these saliency maps using a simple threshold. The red boxes denote ground truth bounding boxes for the associated caption token. Note that saliency maps have been generated for phrases like \textit{a heavy cart full of bricks} and \textit{large group of people}.}
    \label{fig:intro_fig}
\end{figure}

\section{Introduction}

Multi-modal data fusion is critical in retrieving a uniform representation of such data, thereby leading to a better understanding of the underlying semantic relationships \cite{clinchant2011semantic}. It is understood that data from multiple modalities having similar semantic context are characterized by latent relationships. The task of extracting a uniform representation of multimodal data is driven by the hypothesis that these latent relationships can be perceived as meaningful associations when projected to a common space. For instance, \cite{baltruvsaitis2019multimodal} showed that by reasonably projecting data of different modalities into a single common space, one can find a way to reveal several implicit correlation patterns between the different modalities. In real-world situations, upon being introduced to rich textual descriptions of a corresponding image, we can intuitively identify complex relationships between the different visual components and can also localize the different textual entities on the visual scene. Interestingly, we learn these relationships without the level of extensive supervision available in current multi-modal datasets. Consequently, one may be tempted to explore the possibility of developing a method that can learn these beneficial latent relationships naturally in a joint embedding space. We propose a method which is able to find accurate spatial attention maps corresponding to each constituent word or phrase in a descriptive caption associated with an image. These spatial attention maps are derived without any supervision in terms of localizing these textual entities on the image.

We aim to extract relevant inherent phrase localization information from a model trained on a ranking objective. We have chosen the Bidirectinal Image Caption Retrieval task as a proxy for the localization objective. A deep neural network model is trained end to end to minimize a loss objective for this proxy retrieval task\cite{squire2000content}. We hypothesize that by using such a proxy task for our neural network framework, we can implicitly capture visual associations of caption tokens and objects located at different spatial locations in the image. The model does not use any additional parameters, object detection frameworks or region proposal networks to aid the principal localization objective. This localization objective is a favorable outcome of the proxy task training. The retrieval framework that we use resembles some of the well-known retrieval architectures, but has been modified slightly to enable the extraction of the required spatial saliency maps. These saliency maps should represent the implicit localization tendency of a model trained on the retrieval task. We show that a machine can learn to discover visual entities being mentioned as textual segments from a corresponding descriptive caption. Our model can process both words and phrases when finding out these saliency maps.

 Datasets like Flickr30K Entities\cite{plummer2015flickr30K} or Microsoft COCO\cite{lin2014microsoft} have segmentation and ground truth annotation information embedded along with the images and associated captions. This leads to the formulation of a problem where the machine can jointly learn the relationship between language information and visual data while being provided additional localization information. Several methods that depend on such datasets require elaborate and accurate annotations. This often turns out to be a laborious and time intensive tasks. In addition to that, errors in annotations can persist during training and may lead to the development of a fairly wrong or biased embedding space. Our method circumvents this problem by not relying on such bounding box annotations for training.

 In summary, the main contributions of our paper are as follows: We propose a novel framework for localizing query phrases from a caption on its associated image. By using the bidirectional retrieval task as a proxy, our method demonstrates the utility of extracting additional knowledge from an existing network, in a self-supervised fashion. We compare the performance of two different ranking based loss functions and propose the use of the $N$-Pair loss function to optimize the retrieval and localization objective. We also evaluate our framework's localization performance on the Flickr30K Entities dataset and achieve state-of-the-art performance.

\begin{figure*}
    \centering
    \includegraphics[width=2\columnwidth]{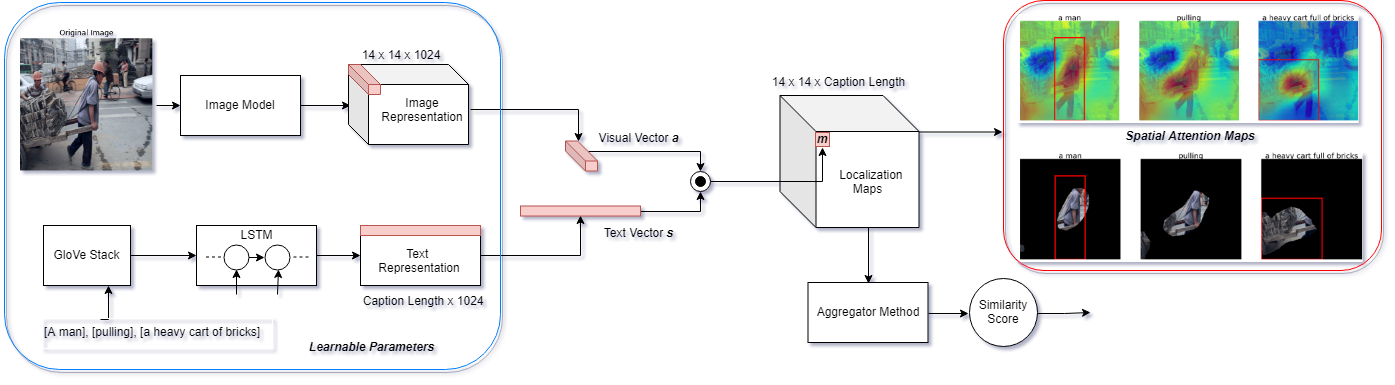}
    \begin{small}

    \caption{Overview of our \textit{Self-Supervised Retrieval Objective} based approach. We use a two branched model architecture that generates an intermediary joint co-localization space and a similarity score across a given caption-image pair based on the score type. A patch in the original image is represented as a feature vector $a_{r,c}$ and a sentence unit is represented as a vector $s_d$, where $r,c$ determine the spatial location of the visual vector, and $d$ determines the position of the sentence token in the caption. A dot product is computed with both these vectors, as highlighted in red, and populates a single position $m_{r,c,d}$ in the localization space.}
    \end{small}
    \label{fig:model_arch}
\end{figure*}

\section{Related Work}

\textbf{Saliency Maps: }Previous work has shown that deep neural networks trained on the image classification task can be used to explore relationships that develop between the network's activation patterns and object labels. However, training neural networks on the image classification tasks that consider a limited number of discrete categories can render the relationships between the objects present in the image to be discrete and disconnected \cite{Frome}. Saliency based methods have been able to use appropriately weighted activation maps derived from the output of a CNN to visualize internal  representations of discrete labels on an image \cite{zhou2016learning}. More recent work has shown that it can also be possible to derive pixel-wise importance for a given class using distinct entity labels\cite{cao2015look}. These methods, however, can not be extended to linguistic structures that represent and convey complex relationships between the visual entities present in the image.

\textbf{Phrase Localization: }The problem of phrase localization has been solved earlier to good effect in a supervised fashion while using large margin based objectives inspired from metric learning \cite{plummer2017phrase, wang2016learning}, but these methods require large amounts of data, while relying heavily on a well-defined set of annotations of object proposals. Moreover, most such methods use a ranking function on a set of region proposals\cite{ren2015faster} to locate regions which best match the textual descriptor. These two steps may not be functionally linked well, as the region proposals may be based on discrete object categories and may not include regions that are relevant to the rich language descriptor. 

A general approach in semi-supervised phrase localization is to optimize a suitably chosen proxy objective. The work in \cite{rohrbach2016grounding} shows that the localization objective can be optimized by reconstructing a phrase correctly after the model learns to attend to a meaningful bounding box. However, this objective can also be optimized by learning better unimodal co-occurrence statistics for language tokens. Xiao et al.\cite{xiao2017weakly} proposed that the task of localization can be optimized by applying a discriminative loss on the whole phrase instead of the object to be localized. Ramanishka et al.\cite{ramanishka2017top} use a caption generation framework to score the phrase on a set of proposal boxes to select the box with highest probability. We argue that the precise reconstruction of a textual descriptor from an image might not correlate well with the localization objective. Several other works  such as \cite{zhang2018top, ramanishka2017top} average over heat-maps to get phrase level localization output. These methods generate attention maps for a single word and need to average over attention maps to find out a single heatmap for a phrase in the caption. The work that most relates to our method is the one presented in \cite{karpathy2014deep}, where the authors use representations of fragmented images and text to obtain a global score. They hypothesize that if matching elements are present in corresponding sentence and image fragments, a fixed non-linear function should generate a high global score. They optimize their objective using a Multiple Instance Learning approach, but the method suffers from the problems arising from using predefined region proposals. A recent work \cite{javed2018learning}uses concept learning as a task to learn self-supervised localization patterns, but can incur the problem of reduced generalization and limitations in the number of possible concepts to be learned. Deep Reinforcement Learning based methods have also been explored \cite{wu2017end} that basically train an agent to move and reshape a bounding box to localize the object according to referring expressions. The model consists of a spatial and a temporal context that affects small changes in predicted bounding boxes at each step which makes it prone to failure in terms of capturing global context.

\section{Model Architecture}

\subsection{Methodology}
The architecture that we develop should be able to construct a high-level joint  embedding space $C$, where instances belonging to a corresponding pair of data from different modalities can be projected close to each other. The model comprises of two different branches to develop two separate representative vectors for the image and caption, respectively. The model receives a batch of corresponding image-caption pairs during training. A caption is sampled randomly from the pool of available captions while constructing the training mini-batch. The corresponding set of images is then retrieved to complete the batch. Each image $I$ in the batch is passed through the image branch of the model to obtain an activation map. Each point on this activation map encodes a certain region on the original image as an $L$-dimensional vector, where $L$ is the depth of the activation block. The network is able to encode $R$ such overlapping regions on the image. The image representation can then be written as  $\textbf{A} = [a_1, a_2, ... , a_R] $. Each caption token $w$ in a single caption $C$ is transformed into an $L$ dimensional vector by the caption branch of the network. This transformation is achieved by passing the GloVE embedding of that token through an LSTM network. These vectors are then stacked one above the other to generate the caption representation $ \textbf{S} = [s_1, s_2, ... , s_N]$. $N$ signifies the length of the caption, or the number of caption tokens considered.

After obtaining a representation for both the image and caption, we use them directly to find a joint associative localization space. The value at each point in this localization space is computed by evaluating the dot product between a caption token vector $s_i$ and a visual vector $a_j$. Consequently, we generate $N$ such maps, and each map consists of $R$ pixels. The weight on each pixel in one of these maps correspond to the degree of association between that specific image region encoded by $a_j$ and the caption token $s_i$. The principal objective at hand for this network is to assign a similarity score to a given image-caption pair. To learn this assignment, one has to train the network to assign a higher score to a similar pair and a lower score to a dissimilar pair. This score is computed by applying an aggregator function on the associative localization space we just built. We use a pooling based method to retrieve a single score from each frame in the joint space.  We should note that each caption token has a corresponding spatial attention map $c_i$ as a part of this joint space, which can be represented as a set of $N$ frames $C = [c_1, c_2, ... ,c_N]$. Therefore, we max-pool across the spatial dimensions for each attention map corresponding to a single token. This gives us a $N$-dimensional vector. We then apply average pooling on this vector to obtain a single scalar value. We name  this aggregator operation the $MaxImage$ operation and can be formulated as: 

{\small
 \begin{equation}
    \mathnormal{S(I,T)_{Max Image} = \frac{1}{n_d}  \sum_{d=1}^{N_d} \max_{r,c}(m_{r,c,d})}
\end{equation}}

Here, $m_{r,c,d}$ represents each point in the 3-dimensional localization space, determined by the row, column and depth dimensions. When using $MaxImage$, the model tries to adjust the spatial location of the maximum activated point for each textual token. This leads to much more refined adjustments as the model picks an image region that correlates highly with the given caption token. It is to be noted that this scalar score is needed to optimize the Bidirectional Retrieval task, which is the primary objective of the network. This score can be computed by a number of aggregator functions, but $MaxImage$ enables us to learn the association between image regions and caption tokens in a much more meaningful fashion.

\subsection{Loss Function}

A ranking based loss function is needed to optimize the chosen proxy task of bidirectional image caption retrieval. For this purpose, we first explored the triplet loss, which uses a triplet of the anchor example itself, a positive sample and an impostor sample. The loss is designed to maximize the similarity between the anchor and the positive instance, and simultaneously minimize the similarity between the anchor and the negative sample. The overall loss $L$, comprises of two components. The first component $L_1$ is derived by fixing an image anchor $I_j$ and finding a pair of captions, comprising of the positive($T_j$) and negative($T_j^{imp}$) samples from the batch. The impostor caption is sampled using an online hard triplet mining strategy. The component $L_2$ uses an anchor caption and a pair of Images($I_j, I^{imp}$) to form the triplet. The overall loss is computed over the entire batch consisting of $N$ samples. 

{\footnotesize
 \begin{equation}
    \mathit{L_1(I_j,T_j,T^{imp}) = max(0, S(I_j, T^{imp})-S(I_j, T_j)+\eta)}
 \end{equation}
 \begin{equation}
    \mathit{L_2(T_j,I_j,I^{imp}) = max(0, S(T_j, I^{imp})-S(T_j, I_j)+\eta)}
 \end{equation}
 \begin{equation}
    \mathit{L = \frac{1}{N}\sum_{j=1}^{N} (L_1(I_j, T_j, T^{imp}) + L_2(T_j, I_j, I^{imp}))}
  \end{equation}
  }

\begin{figure}
    \centering
    \includegraphics[scale=0.4]{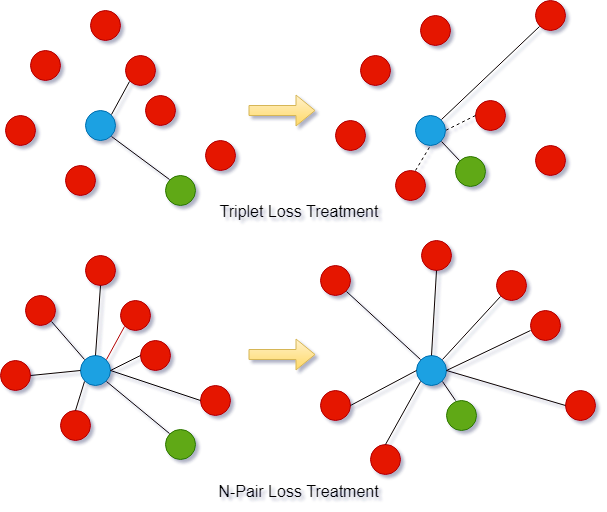}
    \caption{Triplet Loss considers only a single impostor(in red) sample for the anchor(in blue). During a single update, this impostor sample is pushed away from the anchor, while the positive sample(in green) is brought closer. However, other impostor samples, like the ones joined by dotted lines, may still lie close to the anchor. On the other hand, $N$-Pair Loss pushes all negative samples at once, while bringing the positive sample closer.}
    \label{fig:loss_compare}

\end{figure}

After performing several experiments, we found that the Triplet loss converged slowly and the online triplet mining procedure was computationally quite expensive. This led us to explore the possibility of using a much more generalized loss function, like the $N$-Pair loss\cite{sohn2016improved}. Unlike the Triplet loss, the $N$-Pair loss considers all the impostor samples present in a batch. This enhances the discriminative power of the model. During a single update of the triplet loss, the model's parameters are updated based on a single positive and impostor pair, while ignoring the other negative examples in the batch. This leads to a situation where the anchor sample can be distant from a specific impostor class, but might still be considerably close to other impostor samples, as shown in Figure \ref{fig:loss_compare}. The Triplet Loss formulation is based on the assumption that over several iterations, a considerable number of triplets will be sampled, so that the final distance from all impostor samples are greater than the margin $\eta$. This can lead to unstable individual updates and may lead to poor convergence, as observed from our experiments. The $N$-Pair loss formulation is a simple generalization on the triplet loss, so that each update considers multiple impostor samples.

{\footnotesize
 \begin{equation}
    \mathit{L_1(I_j,T_1,...,T_N) = -log(\frac{e^{S(I_j, T_j)^+}}{e^{S(I_j, T_j)^+} + \sum_{i \neq j}^{N}e^{S(I_j, T_j^{imp})}})}
 \end{equation}
 \begin{equation}
    \mathit{L_2(T_j,I_1,...,I_N) = -log(\frac{e^{S(I_j, T_j)^+}}{e^{S(I_j, T_j)^+} + \sum_{i \neq j}^{N}e^{S(I_j^{imp}, T_j)}})}
 \end{equation}
 \begin{equation}
    \mathit{L(I_1,...,I_N,T_1,...,T_N) = \frac{1}{N}\sum_{j=1}^{N} (L_1+ L_2)}
  \end{equation}
  }

We see that the $N$-Pair loss formulation is similar to a Multiclass Softmax structure, and the similarity scores are analogous to the class probability scores in the softmax formulation. During training, the gradient from this loss function ensures that the following update increases the similarity score for the similar pair, and distances all the negative samples from the anchor. This leads to the development of a set of well-structured clusters of similar data points, as seen in Figure \ref{fig:loss_compare}. We also note that in the case where $N = 2$, the $N$-Pair Loss is equivalent to the Triplet Loss. We are the first to propose the use of $N$-Pair Loss in such a setting and it enables us to perform well on the Localization Objective.

\begin{figure}

    \subfloat[\textit{Caption Retrieval from Input Image}]{
        \includegraphics[clip, width=\columnwidth]{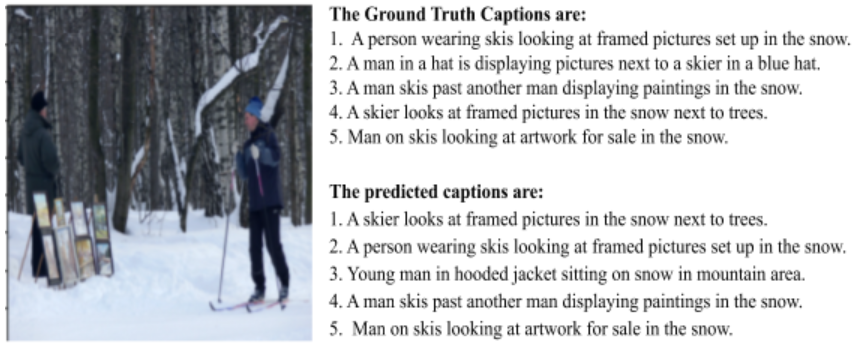}
    }\\
    \subfloat[\textit{Image Retrieval from Input Caption}]{
        \includegraphics[clip, width=\columnwidth]{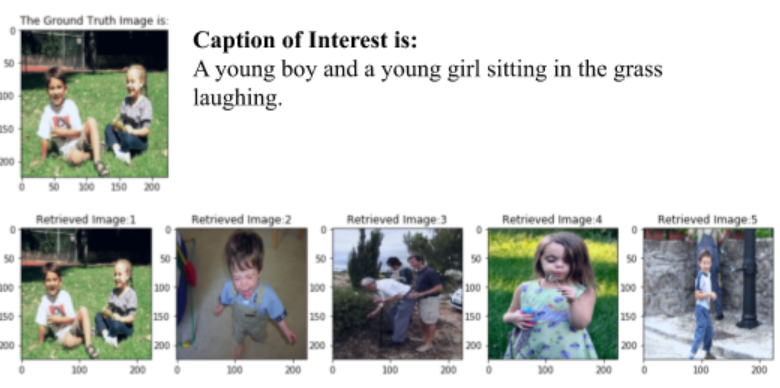}
    }
   \caption{Qualitative Analysis of Bidirectional Retrieval Performance. We notice in (a) that for an input image a closely related set of captions is retrieved, with a considerable intersection over the set of ground truth captions. Similarly, in (b) for an input caption, the model is able to retrieve the ground truth image and another set of images that convey similar semantic context.}
   \label{fig:retrieval_qual}
\end{figure}

However, due to the larger number of objects present per category, as well as the higher number of training images in MSCOCO, we use it for training purposes. The Flickr30K dataset has been used for evaluation purposes.

\section{Experimental Setup}

In this section, we discuss about some Implementational Details, datasets being used, evaluation metrics employed for performance assessment and some baselines.

\subsection{Implementational Details}
The model is trained using the PyTorch framework\cite{ketkar2017introduction}, while using a batch size of 64 and optimized using SGD with momentum. For the image branch, a pretrained VGG-19 model is used to obtain the activation map. The activation map has a dimension of $14$ x $14$ x $1024$. The caption representation has dimensions of $22$ x $1024$. Thus, the associative localization space has dimensions of $14$ x $14$ x $22$. During training, the model is trained end to end. Therefore, weights in the VGG network as well as the LSTM network are updated. Based on some initial analysis, it was found that ($\>95$\%) of captions in both Flickr30K and MSCOCO were of length 20 or lower. Therefore, the pad limit was set to length 20 and only captions of the said length or lower were considered. The development of a batch suitable for the $N$-Pair loss can get computationally expensive. In an ideal scenario, one would like to sample all possible impostor samples for a given anchor. However, we limit the number of impostor samples to $N-1$, as prescribed in the original paper. Under these conditions, we need to develop $2N(N+1)$ pairs in total for a single batch of training. This complexity is reduced by employing a simple workaround. The impostor samples for a given anchor is sampled from the corresponding positive samples for the other anchors in the batch. Therefore, a positive sample for an anchor becomes an impostor sample for another anchor. We avoid sampling captions from the same image to avoid multiple positive samples. This strategy enables us to build a batch from $N$ image-caption pairs, instead of $2N(N+1)$ pairs.

\subsection{Evaluation}

\textbf{Datasets}
The datasets that can be considered for this task have to pair images with global text descriptions and also include some region level annotations for constituent segments in the caption. We choose the MSCOCO \cite{lin2014microsoft} and Flickr30K Entities \cite{plummer2015flickr30K} datasets for this work. Both datasets provide an excellent repository of images and a maximum of 5 corresponding captions for each such image. The MSCOCO Dataset has no region level annotations, and therefore, cannot be used for the quantitative evaluation of the localization task, but has been used for evaluating Retrieval performance. The Flickr30K Entities dataset has been used to evaluate both Retrieval and Localization performance. 

The Flickr30K Entities dataset contains 31,783 images. Each image is associated with 5 captions, with 3.52 query phrases in each caption on average. Each query phrase has 2.3 words on average and these phrases have an average noun count of 1.2. This is a highly desirable scenario when considering the dataset for testing a network on a weakly supervised localization objective. Since the attention maps we build correspond to a single region for localizing entities, it would be beneficial to have a query phrase to point to a single bounding box on the image. This dataset also provides multiple bounding box annotations for different description instances within an image. Other datasets that can be considered are the Visual Genome and ReferIt dataset. Both these datasets have descriptions for regions that are less salient. Due to the nature of the Visual Genome crowdsourcing protocol, its object annotations have a much greater redundancy. For instance, the phrases \textit{A boy wearing a shirt} and \textit{This is a little boy} may be associated with completely different bounding boxes, despite referring to the same person in the image. Moreover, these datsets pair specific objects with short descriptions, rather than pairing images with global descriptions that have segment-wise annotations.  Based on \cite{plummer2015flickr30K}, the Flickr30K Entities dataset is best suited for understanding the different ways by which our mind recognizes visual entities and the most salient relationships amongst them. These factors make Flickr30K Entities the best suited dataset for our task as the proposed method aims to find relationships that build inherently between language and image modalities while being trained on the retrieval objective.

\textbf{Evaluation Metric: }To evaluate the localization maps generated by the model, one needs a fairly suitable evaluation metric. Since the model generates the localization in the form of an attention map, one can use the pointing game metric \cite{zhang2018top}. This metric essentially measures if the most confident region of the predicted attention map falls within the ground truth bounding box. A good attention map can be considered to be consistent if the maximum attention is focused in the ground truth bounding box, which is synonymous to a \textit{Hit} case. In the \textit{Miss} case, it falls outside the ground truth bounding box. The accuracy is given as the ratio of the total number of \textit{Hit} cases to the total testing instances: $\frac{\textit{\#Hit}}{\textit{\#Hit + \#Miss}}$. Many of the previous works have used this metric to generate results. This provides a strong platform to compare this result with state of the art. A well-acknowledged problem with this method arises when there are multiple instances of an entity in the image. The associated textual token will then have multiple bounding boxes on the image. We checked the metric across all such bounding boxes, irrespective of the number of bounding boxes for any given textual token.

\textbf{Baselines: }To compare the proposed method with previous results, a number of suitable baselines have been considered. The first baseline can be a method that chooses the mid-point of each image as the maximum of the pointing game. If the dataset is center-biased, this can provide reasonably good results. Another baseline to be used is to use just a VGG19 model, pretrained on ImageNet data to generate an averaged attention map for the entire image. If this baseline shows high accuracy, the dataset has a bias in which its phrases of the caption are mostly describing the principal object of the image. The accuracy values for these baselines have been provided in \cite{javed2018learning}. Another baseline that has been described here is a method that uses the proposed model, without any training. This is to prove that with improving retrieval performance, the localization performance increases as well. Apart from these baselines, the results here have been compared to the weakly supervised works in \cite{fang2015captions}, \cite{zhang2018top}, \cite{xiao2017weakly}, \cite{ramanishka2017top}, and \cite{javed2018learning}. The compiled results have been presented in Table \ref{table:loc_compare}. We also evaluated the localization performance using the same model, but with the Triplet Loss. Apart from these, cross dataset performance has also been checked by using different trained models for evaluation. Loss types and parse mode evaluation scores are also mentioned in Table \ref{table:loc_expt}. All quantitative evaluation is based on the Flickr30KEntities dataset.

\textbf{Parse Mode:} For a given image-caption pair in the Flickr30K Entities dataset, the spatial attention maps can be extracted for each word in the caption. However, since the dataset associates some visual objects with multi-word phrases within the caption, experiments were performed with two different parse modes. In the default setting,  the spatial attention maps for each constituent word in the descriptive phrase was averaged over to generate a single attention map. In the \textit{phrase-mode} setting, the entire phrase in a given token is represented by a single GloVe vector by averaging over the GloVe vectors for each constituent word in the phrase. This automatically generates a single attention map for the whole phrase. We obtain better results when using the \textit{phrase} parse mode.

\section{Results}

\begin{figure*}

    \subfloat[\textit{Input: A zebra standing in some tall grass by a tree}]{
        \includegraphics[clip, width=2\columnwidth]{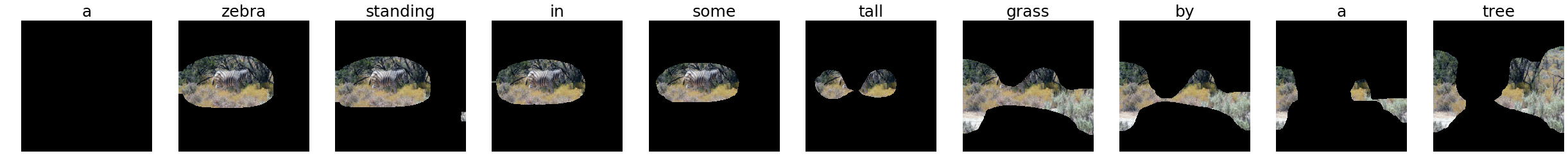}
    }\\
    \subfloat[\textit{Input: Caution sign on a pole in a tropical area with palm trees in the background}]{
        \includegraphics[clip, width=2\columnwidth]{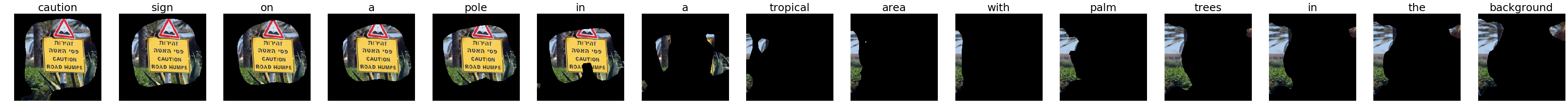}
    }\\
    \subfloat[\textit{Input: A table that has food and a laptop on it}]{
        \includegraphics[clip, width=2\columnwidth]{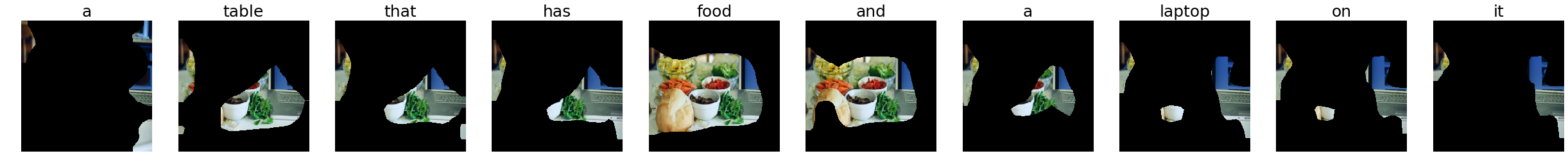}
    }\\
    \subfloat[\textit{Input: A margarita pizza in a restaurant with a soda on the side.}]{
        \includegraphics[clip, width=2\columnwidth]{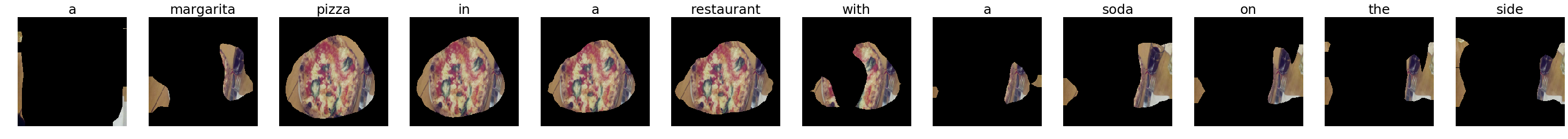}
    }
   \caption{Segmentation Masks based on co-localization maps show the correspondence between caption tokens and objects in the image. A suitable threshold can be used for generating these segmentation masks. These masks enable us to specifically focus only on the region of interest and find out what the machine discovers with each mention of an entity in the caption.}
    \label{fig:seg_masks}
\end{figure*}

\begin{figure*}

    \subfloat[\textit{Input: A public clock in front of a public facility building}]{
        \includegraphics[clip, width=2\columnwidth]{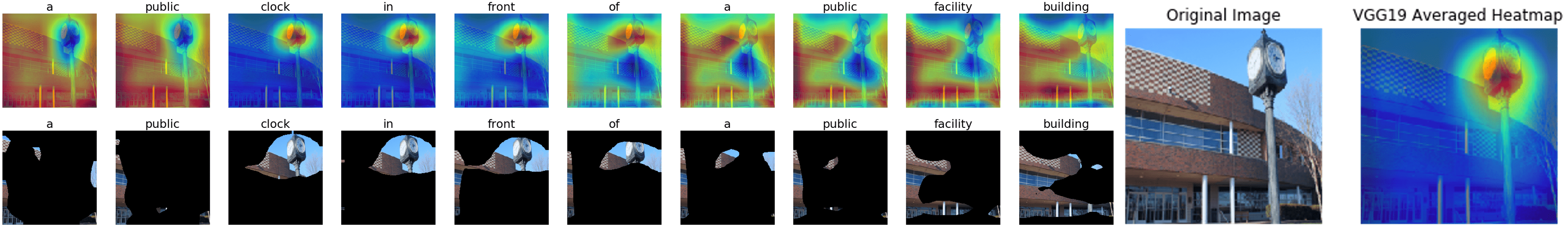}
    }\\
    \subfloat[\textit{Double decker bus decorated with signs is in front of a building with a clocktower}]{
        \includegraphics[clip, width=2\columnwidth]{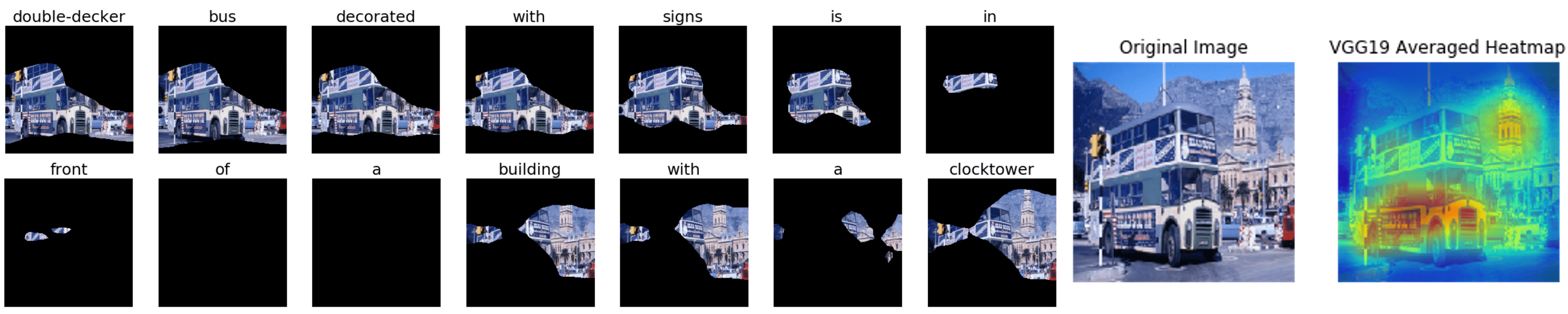}
    }
   \caption{Comparison with VGG19 Averaged Heatmap. Images contain multiple objects and with the occurence of each such entity in the caption, the model is able to focus on different parts of the image. This is clearly distinct from the output of the VGG19 network which focuses mainly on the most salient parts of the image all at once.}
   \label{fig:vgg_19}
\end{figure*}

\subsection{Localization maps}
For a given positive image-caption pair, the intermediary co-localization maps have been extracted and each mask is overlaid on top of the associated image to visualize the most salient parts of the image corresponding to the aligned word in the caption. By setting a threshold on the saliency maps, one can also generate segmentation masks that help us visualize only the most salient parts of the image for a given textual token. It is seen in Figures \ref{fig:intro_fig} and \ref{fig:seg_masks} that the model performs well and is consistent with the initial hypothesis of implicitly generating localization maps as a favorable by-product of the retrieval task. In Figure \ref{fig:intro_fig}, we can also see the heatmap-based visualization of the localization patterns formed on the image for different caption units. Since quantitative evaluation on the MSCOCO dataset is not possible, we present some qualitative results from the MSCOCO test set in Figure \ref{fig:seg_masks}. More qualitative results have been provided in the Supplementary Material.



\begin{table*}[!ht]
\centering
\small
\begin{tabular}{|l|c|c|c|c|c|c|c|c|c|c|c|c|}
\hline
\multicolumn{1}{|c|}{\multirow{3}{*}{Model/Method Name}} & \multicolumn{6}{c|}{Flickr30K Test Set}                                       & \multicolumn{6}{c|}{MSCOCO Test Set}                                          \\ \cline{2-13} 
\multicolumn{1}{|c|}{}                                   & \multicolumn{3}{c|}{Image-to-Caption} & \multicolumn{3}{c|}{Caption-to-Image} & \multicolumn{3}{c|}{Image-to-Caption} & \multicolumn{3}{c|}{Caption-to-Image} \\ \cline{2-13} 
\multicolumn{1}{|c|}{}                                   & R@1         & R@5        & R@10       & R@1         & R@5        & R@10       & R@1         & R@5        & R@10       & R@1         & R@5        & R@10       \\ \hline
mCNN(ensemble)\cite{ma2015multimodal}                                           & 33.6        & 64.1       & 74.9       & 26.2        & 56.3       & 69.6       & 42.8        & 73.1       & 84.1       & 32.6        & 68.6       & 82.8       \\ \hline
m-RNN-VGG\cite{mao2014deep}                                             & 35.4        & 63.8       & 73.7       & 22.8        & 50.7       & 63.1       & 41.0        & 73.0       & 83.5       & 29.0        & 42.2       & 77.0       \\ \hline
CCA (Fisher Vector)\cite{klein2014fisher}                                     & 35.0        & 62.0       & 73.8       & 25.0        & 52.7       & 66.0       & 39.4        & 67.9       & 80.9       & 25.1        & 59.8       & 76.6       \\ \hline
DSPE (Fisher Vector)\cite{wang2016learning}                                 & 40.3        & 68.9       & 79.9       & 29.7        & 60.1       & 72.1       & 50.1        & 79.7       & 89.2       & 39.6        & 75.2       & 86.9       \\ \hline
Proposed Method N-Pair Loss                              & 27.0        & 49.0       & 62.0       & 10.0        & 32.0       & 42.0       & 47.0        & 77.0       & 92.9       & 27.9        & 65.9       & 81.0       \\ \hline
\end{tabular}
\caption{Quantitative Evaluation of Bidirectional Image Caption Retrieval task for both datasets. Recall scores are computed as Recall@1, Recall@5 and Recall@10. The scores do not better the state of the art results in the proxy retrieval task, but the model performs well in the primary localization objective. Recall scores are considerably poorer for the model using Triplet Loss in our method.}

\label{table:retrieval}
\end{table*}

\begin{table}[!ht]
\centering
\begin{tabular}{|l|l|}
\hline
\textbf{Method}                                                                       & \textbf{Accuracy} \\ \hline
Random Baseline                                                                       & 27.24             \\ \hline
Center Baseline                                                                       & 49.20             \\ \hline
VGG Baseline                                                                          & 35.37             \\ \hline
No training Baseline                                                                  & 26.40             \\ \hline
Fang \textit{et al} \cite{fang2015captions}        & 29.03             \\ \hline
Zhang \textit{et al} \cite{zhang2018top}           & 42.40             \\ \hline
Ramanishka \textit{et al} \cite{ramanishka2017top} & 50.10             \\ \hline
Javed \textit{et al} \cite{javed2018learning} & 49.10             \\ \hline

\textbf{Proposed Method - $N$-Pair Loss}                                                     & \textbf{51.06}    \\ \hline
Proposed Method - Triplet loss&14.93 \\ \hline
\end{tabular}
\vspace{0.1cm}
\caption{Quantitative Evaluation of Various State of the Art methods and proposed method. Several Baselines have also been shown for suitable comparison. Tests have been perormed on Flickr30K Entities Test Fold. It is seen that our method surpasses all state of the art results.}
\label{table:loc_compare}
\end{table}

\textbf{Quantitative Evaluation: }Table \ref{table:loc_compare} presents results for all baselines, previous methods and proposed method for the Flickr30K Test dataset. It is seen that this method clearly outperforms the other state of the art methods by a fair margin. Based on the baseline analysis, an interesting takeaway is the fact that the Flickr30K Entities dataset is indeed slightly biased towards the center point and a lot of phrases do have their bounding boxes encompassing the central point in the image. This proposed method is able to surpass the results in \cite{ramanishka2017top}, which uses a captioning network and an attention mechanism to generate such attention maps. It is also found that the model performs fairly well on the localization objective across datasets. We trained a model on the MSCOCO Dataset and evaluated the localization objective on the Flickr30K Entities dataset and found that in the Phrase Parse Mode, this model trained on the MSCOCO dataset performs almost as well as the model trained on Flickr30K Entities dataset with the default Phrase mode. We also note that the $N$-Pair Loss formulation helps us achieve much better results when compared to the Triplet Loss formulation. Results from further experiments have been included in the Supplementary Material. 

\textbf{Comparison with VGG19 Baseline: }To have a fair comparison, we also compare our results qualitatively and quantitatively with a VGG19 Baseline to ensure that our model is learning an additional localization objective. The quantitative evaluation has been presented in Table \ref{table:loc_compare} and shows that the model indeed is able to distinguish different parts of the images containing several objects being mentioned in the associated caption. Looking at the results in Figure \ref{fig:vgg_19}, we can see that the VGG19 Averaged heatmaps provide a single region of focus on the most salient portion of the image. Whereas, this proposed model intelligently looks at different parts of the image for distinct mentions of the various entities in the caption. This substantiates the hypothesis that a retrieval framework inherently learns to discover different visual objects in the image as it learns to associate similar text and image data.

\subsection{Retrieval Scores}
Retrieval scores are reported as Recall@1, Recall@5 and Recall@10 for bidirectional retrieval tasks. The model seems to perform decently well on the test fold of Flickr30K and a separate test fold held out from the MSCOCO Validation set. A qualitative evaluation, presented in Figure \ref{fig:retrieval_qual}, shows how the model extracts one of the ground truth instances being retrieved from the ranked list. Table \ref{table:retrieval} lists all the recall scores for models using different score types on an image fold of size 100. As the standard model for the retrieval task has been slightly altered, it was expected that the retrieval performance would be a bit worse than the state of the art results. But, experiments showed that as the retrieval performance increased, so did the localization performance. Moreover, it being a proxy task, we focused more on tweaking the model to get better localization performance. As expected Retireval Performance was very poor on the model trained using Triplet Loss.

\section{Conclusion}
We propose a novel neural architecture that can find localization patterns across the text and image modalities. After several experiments, this architecture shows improved performance on the Flickr30K Entities dataset when compared to other state of the art methods in self-supervised localization. The method leverages the inherent localization occuring across sentence units and the image when being trained for a bidirectional retrieval objective. This is critical in ensuring that there are no excess parameters or weights involved in obtaining these maps. Further experimentations enabled the design of a better optimization technique by using the $N$-Pair loss function. However, loss metrics like the one mentioned in \cite{oh2017deep} can be explored, to learn the semantic relationships better. This experiment also proves that the machine is able to localize the visual region corresponding to a specific token in the textual caption, and can also discover relevant visual regions on its own without any form of supervision.

{\small
\bibliographystyle{ieee}
\bibliography{egbib}

\begin{thebibliography}{10}\itemsep=-1pt

\bibitem{baltruvsaitis2019multimodal}
T.~Baltru{\v{s}}aitis, C.~Ahuja, and L.-P. Morency.
\newblock Multimodal machine learning: A survey and taxonomy.
\newblock {\em IEEE Transactions on Pattern Analysis and Machine Intelligence},
  41(2):423--443, 2019.

\bibitem{cao2015look}
C.~Cao, X.~Liu, Y.~Yang, Y.~Yu, J.~Wang, Z.~Wang, Y.~Huang, L.~Wang, C.~Huang,
  W.~Xu, et~al.
\newblock Look and think twice: Capturing top-down visual attention with
  feedback convolutional neural networks.
\newblock In {\em Proceedings of the IEEE International Conference on Computer
  Vision}, pages 2956--2964, 2015.

\bibitem{clinchant2011semantic}
S.~Clinchant, J.~Ah-Pine, and G.~Csurka.
\newblock Semantic combination of textual and visual information in multimedia
  retrieval.
\newblock In {\em Proceedings of the 1st ACM international conference on
  multimedia retrieval}, page~44. ACM, 2011.

\bibitem{fang2015captions}
H.~Fang, S.~Gupta, F.~Iandola, R.~K. Srivastava, L.~Deng, P.~Doll{\'a}r,
  J.~Gao, X.~He, M.~Mitchell, J.~C. Platt, et~al.
\newblock From captions to visual concepts and back.
\newblock In {\em Proceedings of the IEEE conference on computer vision and
  pattern recognition}, pages 1473--1482, 2015.

\bibitem{Frome}
A.~Frome, G.~S. Corrado, J.~Shlens, S.~{Bengio Jeffrey Dean}, A.~Ranzato, and
  T.~Mikolov.
\newblock {DeViSE: A Deep Visual-Semantic Embedding Model}.
\newblock Technical report.

\bibitem{javed2018learning}
S.~A. Javed, S.~Saxena, and V.~Gandhi.
\newblock Learning unsupervised visual grounding through semantic
  self-supervision.
\newblock {\em arXiv preprint arXiv:1803.06506}, 2018.

\bibitem{karpathy2014deep}
A.~Karpathy, A.~Joulin, and L.~F. Fei-Fei.
\newblock Deep fragment embeddings for bidirectional image sentence mapping.
\newblock In {\em Advances in neural information processing systems}, pages
  1889--1897, 2014.

\bibitem{ketkar2017introduction}
N.~Ketkar.
\newblock Introduction to pytorch.
\newblock In {\em Deep learning with python}, pages 195--208. Springer, 2017.

\bibitem{klein2014fisher}
B.~Klein, G.~Lev, G.~Sadeh, and L.~Wolf.
\newblock Fisher vectors derived from hybrid gaussian-laplacian mixture models
  for image annotation.
\newblock {\em arXiv preprint arXiv:1411.7399}, 2014.

\bibitem{lin2014microsoft}
T.-Y. Lin, M.~Maire, S.~Belongie, J.~Hays, P.~Perona, D.~Ramanan,
  P.~Doll{\'a}r, and C.~L. Zitnick.
\newblock Microsoft coco: Common objects in context.
\newblock In {\em European conference on computer vision}, pages 740--755.
  Springer, 2014.

\bibitem{ma2015multimodal}
L.~Ma, Z.~Lu, L.~Shang, and H.~Li.
\newblock Multimodal convolutional neural networks for matching image and
  sentence.
\newblock In {\em Proceedings of the IEEE international conference on computer
  vision}, pages 2623--2631, 2015.

\bibitem{mao2014deep}
J.~Mao, W.~Xu, Y.~Yang, J.~Wang, Z.~Huang, and A.~Yuille.
\newblock Deep captioning with multimodal recurrent neural networks (m-rnn).
\newblock {\em arXiv preprint arXiv:1412.6632}, 2014.

\bibitem{oh2017deep}
H.~Oh~Song, S.~Jegelka, V.~Rathod, and K.~Murphy.
\newblock Deep metric learning via facility location.
\newblock In {\em Proceedings of the IEEE Conference on Computer Vision and
  Pattern Recognition}, pages 5382--5390, 2017.

\bibitem{plummer2017phrase}
B.~A. Plummer, A.~Mallya, C.~M. Cervantes, J.~Hockenmaier, and S.~Lazebnik.
\newblock Phrase localization and visual relationship detection with
  comprehensive image-language cues.
\newblock In {\em Proceedings of the IEEE International Conference on Computer
  Vision}, pages 1928--1937, 2017.

\bibitem{plummer2015flickr30K}
B.~A. Plummer, L.~Wang, C.~M. Cervantes, J.~C. Caicedo, J.~Hockenmaier, and
  S.~Lazebnik.
\newblock Flickr30k entities: Collecting region-to-phrase correspondences for
  richer image-to-sentence models.
\newblock In {\em Proceedings of the IEEE international conference on computer
  vision}, pages 2641--2649, 2015.

\bibitem{ramanishka2017top}
V.~Ramanishka, A.~Das, J.~Zhang, and K.~Saenko.
\newblock Top-down visual saliency guided by captions.
\newblock In {\em Proceedings of the IEEE Conference on Computer Vision and
  Pattern Recognition}, pages 7206--7215, 2017.

\bibitem{rohrbach2016grounding}
A.~Rohrbach, M.~Rohrbach, R.~Hu, T.~Darrell, and B.~Schiele.
\newblock Grounding of textual phrases in images by reconstruction.
\newblock In {\em European Conference on Computer Vision}, pages 817--834.
  Springer, 2016.

\bibitem{sohn2016improved}
K.~Sohn.
\newblock Improved deep metric learning with multi-class n-pair loss objective.
\newblock In {\em Advances in Neural Information Processing Systems}, pages
  1857--1865, 2016.

\bibitem{squire2000content}
D.~M. Squire, W.~M{\"u}ller, H.~M{\"u}ller, and T.~Pun.
\newblock Content-based query of image databases: inspirations from text
  retrieval.
\newblock {\em Pattern Recognition Letters}, 21(13-14):1193--1198, 2000.

\bibitem{wang2016learning}
L.~Wang, Y.~Li, and S.~Lazebnik.
\newblock Learning deep structure-preserving image-text embeddings.
\newblock In {\em Proceedings of the IEEE conference on computer vision and
  pattern recognition}, pages 5005--5013, 2016.

\bibitem{wu2017end}
F.~Wu, Z.~Xu, and Y.~Yang.
\newblock An end-to-end approach to natural language object retrieval via
  context-aware deep reinforcement learning.
\newblock {\em arXiv preprint arXiv:1703.07579}, 2017.

\bibitem{xiao2017weakly}
F.~Xiao, L.~Sigal, and Y.~Jae~Lee.
\newblock Weakly-supervised visual grounding of phrases with linguistic
  structures.
\newblock In {\em Proceedings of the IEEE Conference on Computer Vision and
  Pattern Recognition}, pages 5945--5954, 2017.

\bibitem{zhang2018top}
J.~Zhang, S.~A. Bargal, Z.~Lin, J.~Brandt, X.~Shen, and S.~Sclaroff.
\newblock Top-down neural attention by excitation backprop.
\newblock {\em International Journal of Computer Vision}, 126(10):1084--1102,
  2018.

\bibitem{zhou2016learning}
B.~Zhou, A.~Khosla, A.~Lapedriza, A.~Oliva, and A.~Torralba.
\newblock Learning deep features for discriminative localization.
\newblock In {\em Proceedings of the IEEE conference on computer vision and
  pattern recognition}, pages 2921--2929, 2016.

\end{thebibliography}
}

\end{document}